# BanglaNirTox: A Large-scale Parallel Corpus for Explainable AI in Bengali Text Detoxification

**Ayesha Afroza Mohsin\*, Mashrur Ahsan\*, Nafisa Maliyat\*, Shanta Maria\*,
Syed Rifat Raiyan, Hasan Mahmud, Md Kamrul Hasan**

Systems and Software Lab (SSL)
Department of Computer Science and Engineering
Islamic University of Technology, Gazipur, Bangladesh
`{ayeshaafroza, mashrurahsan, nafisamaliyat, shantamaria, rifatraiyan, hasan, hasank}`
`@iut-dhaka.edu`

**Abstract**

Toxic language in Bengali remains prevalent, especially in online environments, with few effective precautions against it. Although text detoxification has seen progress in high-resource languages, Bengali remains underexplored due to limited resources. In this paper, we propose a novel pipeline for Bengali text detoxification that combines Pareto class-optimized large language models (LLMs) and Chain-of-Thought (CoT) prompting to generate detoxified sentences. To support this effort, we construct BanglaNirTox, an artificially generated parallel corpus of 68,041 toxic Bengali sentences with class-wise toxicity labels, reasonings, and detoxified paraphrases, using Pareto-optimized LLMs evaluated on random samples. The resulting BanglaNirTox dataset is used to fine-tune language models to produce better detoxified versions of Bengali sentences. Our findings show that Pareto-optimized LLMs with CoT prompting significantly enhance the quality and consistency of Bengali text detoxification.

**Warning:** *This paper contains examples of toxic and offensive language.*

## 1. Introduction

Online Bengali communities face similar challenges with toxic and abusive language as communities in other languages (Das et al., 2021). One promising solution to mitigate this issue without limiting user interaction is *text detoxification*—the task of transforming toxic text into non-toxic form while preserving its original meaning (Khondaker et al., 2024).

Although significant progress has been made in high-resource languages such as English (Logacheva et al., 2022), no prior work has addressed this problem in the context of Bengali. To bridge this research gap, we propose a novel pipeline for generating data while minimizing human annotation efforts. Using this pipeline, we construct BanglaNirTox, a parallel corpus consisting of 68,041 samples with the following features: (i) toxic text, (ii) a rationale explaining the toxicity classification, (iii) the classification (Explicit, Implicit, Not Toxic), (iv) a detoxified version of the original toxic input and (v) an indication of whether the toxic and detoxified texts are paraphrases of each other. Examples of the classification categories can be seen in Table 1.

Furthermore, we fine-tuned two Bengali language models, BanglaLLaMA and BanglaMT5, on this dataset, achieving strong performance on the detoxification task, as evident in Table 2.

| Toxic Text | Classification | Detoxification |
|---|---|---|
| **** **রা ধর্ম নিয়া ব্যবসা করে, (\*\*\*\*er make a business out of religion,) | Explicit | ওরা ধর্মের নামে ব্যবসা করে, (They make a business out of religion,) |
| ভাই এগুলারে পাবনার মানসিক হাসপাতালে ভর্তি করানো উচিত (Bro, these people should be admitted to the Pabna mental hospital) | Implicit | ভাই এনাদের চিন্তাভাবনা এবং কাজকর্ম খুবই অযৌক্তিক। (Brother, their thoughts and actions are very illogical) |
| রেট কত? (What's the rate?) | Not Toxic | – |

Table 1: Examples of the 3 classification categories

Our contributions are as follows:

1. We propose a novel data generation pipeline utilizing Pareto optimality to minimize human annotation efforts.

---
\*These authors contributed equally to this work.

2. We present BANGLANIRTOX, the first Bengali parallel corpus of toxic and detoxified sentence pairs, along with their toxicity rationales.

3. We develop a detoxification module capable of detecting and transforming toxic Bengali text, addressing both implicit and explicit toxicity.

## 2. Literature Review

Text detoxification is a subtask of text style transfer that aims to transform toxic or offensive text into a more neutral or non-toxic form, while preserving the original meaning (Mukherjee et al., 2023). Over the years, style transfer tasks in NLP have been widely explored due to their relevance in applications such as sentiment transfer, politeness conversion, and formality adjustment.

**Text Detoxification.** Traditionally, text detoxification has been approached using non-parallel corpora (Tran et al., 2020; Laugier et al., 2021; Atwell et al., 2022), often relying on masking (Hallinan et al., 2023) or lexicon-based (Sazzed, 2021) techniques to remove toxic content.

**Parallel Corpora.** A significant advancement in this area was the introduction of ParaDetox (Logacheva et al., 2022), which presented the first parallel corpus specifically curated for English text detoxification. This enabled more structured and effective learning for style transfer in this domain. Building on this, MultiParaDetox (Dementieva et al., 2024) extended the framework to multiple languages, offering a multilingual parallel corpus for detoxification tasks. However, this work (Dementieva et al., 2024) did not include Bengali, leaving that linguistic domain largely unexplored.

**Use of LLMs in Text Detoxification.** Both ParaDetox (Logacheva et al., 2022) and MultiParaDetox (Dementieva et al., 2024) relied on human annotation for data labeling, making the creation of such datasets resource-intensive. To address this limitation, (Moskovskiy et al., 2024) proposed leveraging Large Language Models (LLMs) to generate synthetic parallel data for detoxification, thereby significantly reducing the reliance on manual annotation and improving the scalability of dataset creation. (Dementieva et al., 2025) addressed the need for detoxification for multiple languages with the help of explainable LLMs—performing detoxification on multiple languages.

**Detoxified Text Generation.** Most of the previous work focuses on the detection and classification (Belal et al., 2023) of explicit toxicity and then takes further steps to remove the detected toxicity (Dementieva et al., 2021). The detection and mitigation of implicit toxicity is also an important challenge. It involves subtle insults, offensive sarcastic remarks, cultural references that most of the time evade traditional detection methods (Gunturi et al., 2023) (Hartvigsen et al., 2022).

**Text Detoxification in Bengali.** Existing studies have mainly focused on toxicity and hate speech detection rather than the task of detoxifying text. For instance, (Belal et al., 2023) proposed a framework for multi-labeled classification of toxicity. (Sazzed, 2021) contributed a manually curated lexicon, and similarly, ToxLex_bn (Rashid, 2022) provided a dataset specifically targeting Bengali online toxicity. BD-SHS (Romim et al., 2022) offered a comprehensive dataset annotated across multiple social dimensions. Despite these contributions, a significant research gap still exists in the field of Bengali text detoxification.

## 3. Methodology

Figure 1 presents an overview of our proposed pipeline.

### 3.1. Dataset Accumulation and Preparation

We take toxic samples from multiple publicly available Bengali datasets: (i) Multi Labeled Toxic Comments (Belal et al., 2023), (ii) BD-SHS (Romim et al., 2022), (iii) Bangla Hate Speech v1.0 and v2.0. (Karim et al., 2020).

All datasets are standardized to a binary toxicity format and the dataset was preprocessed to remove inconsistencies and improve overall quality before generation.

### 3.2. Pareto Optimality-Based LLM Selection

We adopt a multi-objective evaluation framework which uses the concept of Pareto optimality (Duh et al., 2012). We manually annotate a randomly selected subset of 1,000 instances as "Explicit", "Implicit", or "Not Toxic", following an established annotation guidelines.[1] Due to class imbalance, we manually build a class-balanced subset from the annotated samples by selecting examples from underrepresented categories.

---
[1] Additional Information regarding the annotation guidelines will be delineated in the Appendix section of the camera-ready version.

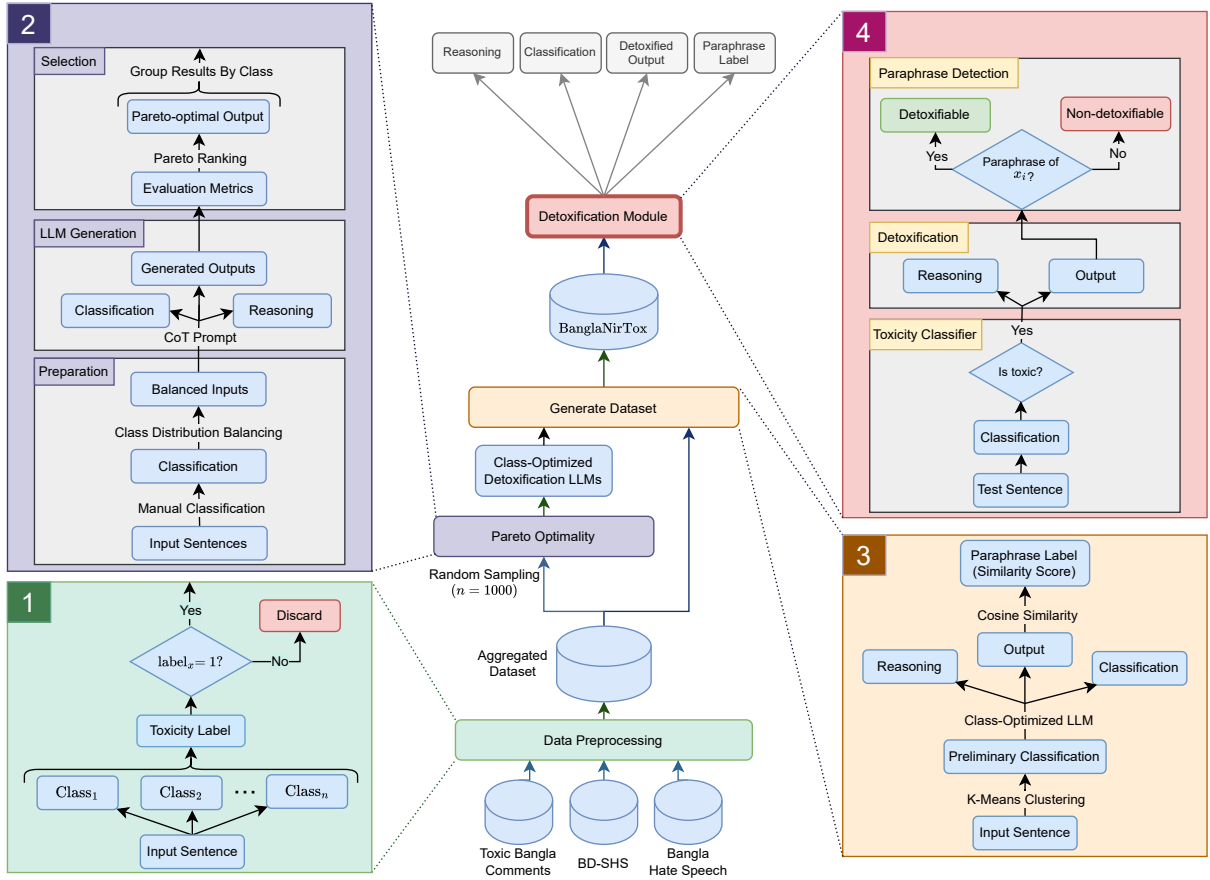

Figure 1: Overview of our methodology comprising four stages: (1) Preprocessing raw data, (2) Generating and ranking detoxified outputs from three LLMs using Chain-of-Thought and Pareto frontier analysis to select the best model, (3) Applying $k$-Means clustering for class-specific detoxification, and (4) Training a final module for reasoning-based classification, detoxification, and paraphrase validation.

**Prompting and Output Generation.** We prompt three candidate LLMs: Grok-3, GPT-4o, and Gemini-2.5 Pro using a Chain-of-Thought (CoT) prompt[2] that guides them to: (1) generate a 15-word Bengali reasoning for toxicity classification, (2) assign a toxicity label, and (3) produce three unique detoxified paraphrases of the original input sentence. Each model is first prompted to generate a concise 15-word reasoning, which ensures a thinking step before the model assigns a classification label, while also limiting the length of the model's response.

We then plot a Pareto frontier over six automated metrics – Style Transfer Accuracy (STA), Semantic Similarity (SIM), Fluency (FL), ROUGE-1, BERTScore, and METEOR – identifying for each input the output that has the least quality trade-offs, *i.e.*, no other output is strictly better across all dimensions. Each output is ranked based on its position, and the language model contributing the greatest number of top-ranked outputs on the Pareto frontier is selected for generating the complete parallel corpus.

### 3.3. Parallel Data Generation Using Class-Optimized LLMs

Our Pareto frontier analysis identified Gemini-2.5 Pro as the most effective LLM across all three toxicity classes. Consequently, we selected Gemini for generating the full parallel dataset. Generation was conducted using another tailored prompt. One limitation encountered during this process was the occasional failure of the API to return outputs for highly toxic inputs, particularly within the Explicit class. The resulting parallel dataset was evaluated using seven metrics (Style Transfer Accuracy (STA), Semantic Similarity (SIM), Fluency (FL), J Score, BERTScore, ROUGE-1, and METEOR).

### 3.4. Detoxification Module

For the detoxification task, we fine-tune several sequence-to-sequence (Seq2Seq) and Large Lan-guage models, including TigerLLM (Raihan and Zampieri, 2025), condBERT (Atwell et al., 2022), mBART (Chipman et al., 2022), BanglaT5 (Bhattacharjee et al., 2022), and BanglaLLaMA (Zehady et al., 2024), using our parallel corpus.

---
[2] Additional Information regarding all the prompts used will be delineated in the Appendix section of the camera-ready version.

Model performance is evaluated the seven metrics shown in Table 2.

## 4. Experimental Setup

After constructing BANGLANIRTOX, we take several baseline models and fine-tune several language models on our dataset. The models are evaluated using seven metrics – Style Transfer Accuracy (STA), Semantic Similarity (SIM), Fluency (FL), J Score, BERTScore, ROUGE-1, and METEOR

**Duplication.** The duplication baseline simply reproduces the input text without modification. As a result, it achieves the highest SIM, BERTScore, ROUGE, and METEOR values. However, this serves only as a control baseline since it performs no detoxification.

**Deletion.** For the deletion baseline, toxic tokens are removed according to the publicly available Bengali toxic lexicon, ToxLex_bn (Rashid, 2022).

**Fine-tuned LM on Translated Data.** We adapt the mBART baseline from the ParaDetox paper (Logacheva et al., 2022). Specifically, the mBART model is first trained on the ParaDetox dataset translated into Bengali using the Google Translation API. The model is then further fine-tuned using the multilingual mBART model (Tang et al., 2020).

**Fine-tuned LM on BANGLANIRTOX.** We further fine-tune several language models directly on our dataset, BANGLANIRTOX. The models include mBART, TigerLLM[3], and BanglaLlama.

**Zero-shot Prompting.** We also experiment with prompting mBART and TigerLLM models in a zero-shot setup. The prompt in this configuration is a simple instruction directing the model to detoxify the given text.

**condBERT Modification.** Since the original condBERT architecture is not suitable for Bengali text, we modify its structure to use the BanglaBERT tokenizer (Kowsher et al., 2022) for tokenization, and replace the base model with BanglaBERT.

---

[3] https://huggingface.co/md-nishat-008/TigerLLM-9B-it

## 5. Results

Evaluation results reveal distinct patterns across model families. While the Duplication and Deletion baselines excel in content preservation, they fail to adequately detoxify text. Models adapted from other languages show mixed efficacy; the Google-translated variant performed reasonably well, potentially because the translation process itself neutralizes some toxicity, whereas the mBART-translated model severely over-corrects the text. The condBERT model struggles with fluency, and zero-shot models show only a nascent detoxification ability. Ultimately, models fine-tuned directly on the BANGLANIRTOX corpus prove to be on the more robust end of the spectrum, demonstrating a superior balance between detoxification, fluency, and meaning preservation, which underscores the critical importance of in-domain, language-specific data.

## 6. Conclusion

We present BANGLANIRTOX, the first parallel corpus of toxic Bengali sentences with class-wise toxicity labels, a concise reasoning for the classification, and the corresponding detoxified equivalent. We also propose a novel pipeline to synthetically generate detoxified equivalents as well as a detoxification module that detects and detoxifies toxicity in Bengali text samples. We evaluate our dataset and also evaluate several baselines against models fine-tuned on BANGLANIRTOX, by using seven metrics (Style Transfer Accuracy (STA), Semantic Similarity (SIM), Fluency (FL), J Score, BERTScore, ROUGE-1, and METEOR) and find that fine-tuned models — particularly BanglaLLaMA, mBART, and TigerLLM — achieve the best balance across all evaluation metrics.

## 7. Limitations

While our work makes progress on Bengali text detoxification, we acknowledge that our work remains impeded by several limitations as follows:

- **Subjectivity of toxicity:** The perception of what constitutes toxic or offensive language is inherently subjective and culturally dependent. This is especially true for implicitly toxic text that depends heavily on context and cultural expression. Expressions that one community may consider offensive might be interpreted neutrally or humorously by another, making a universal labeling challenging to implement.

| Model | STA ↑ | SIM ↑ | FL ↑ | J ↑ | BERT ↑ | ROUGE ↑ | METEOR ↑ |
|---|---|---|---|---|---|---|---|
| Duplication | 0.37 | 1.00 ± 0.0 | 0.50 ± 0.0 | 0.32 ± 0.2 | 1.00 ± 0.0 | 1.00 ± 0.0 | 1.00 ± 0.0 |
| Deletion | 0.44 | 0.95 ± 0.1 | 0.48 ± 0.0 | 0.27 ± 0.2 | 0.96 ± 0.1 | 0.96 ± 0.1 | 0.96 ± 0.1 |
| mBART - ParaDetox Google translated | 0.71 | 0.95 ± 0.1 | 0.52 ± 0.1 | 0.35 ± 0.1 | 0.94 ± 0.1 | 0.93 ± 0.1 | 0.89 ± 0.2 |
| mBART - ParaDetox mBART translated | 0.99 | 0.15 ± 0.1 | 0.85 ± 0.1 | 0.12 ± 0.1 | 0.26 ± 0.1 | 0.01 ± 0.1 | 0.01 ± 0.1 |
| mBART - BanglaNirTox Fine-Tuned | 0.52 | 0.66 ± 0.3 | 0.64 ± 0.1 | 0.21 ± 0.1 | 0.64 ± 0.2 | 0.54 ± 0.3 | 0.43 ± 0.3 |
| mBART - BanglaNirTox (Zero-Shot) | 0.45 | 0.63 ± 0.3 | 0.52 ± 0.1 | 0.14 ± 0.1 | 0.80 ± 0.2 | 0.63 ± 0.3 | 0.68 ± 0.3 |
| condBERT - BanglaNirTox | 0.33 | 0.78 ± 0.1 | 0.29 ± 0.1 | 0.08 ± 0.0 | 0.93 ± 0.1 | 0.54 ± 0.2 | 0.45 ± 0.2 |
| TigerLLM - BanglaNirTox Fine-Tuned | 0.53 | 0.57 ± 0.2 | 0.71 ± 0.1 | 0.21 ± 0.1 | 0.52 ± 0.2 | 0.33 ± 0.2 | 0.26 ± 0.2 |
| TigerLLM - BanglaNirTox (Zero-Shot) | 0.50 | 0.40 ± 0.2 | 0.72 ± 0.1 | 0.14 ± 0.1 | 0.48 ± 0.1 | 0.20 ± 0.1 | 0.29 ± 0.2 |
| BanglaLlama - BanglaNirTox Fine-Tuned | 0.58 | 0.57 ± 0.2 | 0.68 ± 0.1 | 0.22 ± 0.1 | 0.53 ± 0.2 | 0.34 ± 0.2 | 0.23 ± 0.2 |

Table 2: Performance of different models on the BANGLANIRTOX dataset across multiple evaluation metrics

- **Partial manual validation:** The synthetically generated dataset was not exhaustively verified by human annotators. Despite extensive efforts, the resulting dataset may not be as comprehensive as it might have been with human annotators.

- **LLM resource constraints:** The study utilized the most capable Large Language Models that were accessible within the available computational and financial resources. More advanced or larger-scale models could not be employed due to cost and hardware limitations.

- **Limitations of automated metrics:** Automatic toxicity classifiers and evaluation metrics do not fully capture the nuanced effectiveness of detoxification models compared to human judgment. These metrics often fail to detect implicit or context-dependent toxicity. This is especially true for an underrepresented language like Bengali.

## 8. Bibliographical References

## 9. Language Resource References